\def\ritaArxivVersion{}
\documentclass{article}
\PassOptionsToPackage{dvipsnames,table}{xcolor}
\usepackage{iclr2027_conference,times}
\usepackage[T1]{fontenc}
\usepackage{iftex}
\ifPDFTeX
  \usepackage[utf8]{inputenc}
\fi
\usepackage[dvipsnames,table]{xcolor}
\usepackage{graphicx}
\usepackage{booktabs}
\usepackage{makecell}
\usepackage{multirow}
\usepackage{mathtools,amssymb,amsthm}
\usepackage{siunitx}
\usepackage{wrapfig}
\usepackage{needspace}
\usepackage{subcaption}
\usepackage[normalem]{ulem}
\usepackage{soul}
\usepackage{xspace}
\usepackage{url}
\usepackage[hidelinks]{hyperref}

\newif\ifshowedits
\showeditsfalse
\ifshowedits
  \DeclareRobustCommand{\AY}[1]{\textbf{\textcolor{blue}{[AY: #1]}}}
  \DeclareRobustCommand{\pz}[1]{\textbf{\textcolor{magenta}{[PZ: #1]}}}
\else
  \DeclareRobustCommand{\AY}[1]{}
  \DeclareRobustCommand{\pz}[1]{}
  
  \renewcommand{\st}[1]{}
\fi

\theoremstyle{definition}

\theoremstyle{plain}

\makeatletter
\DeclareRobustCommand\onedot{\futurelet\@let@token\@onedot}
\def\@onedot{\ifx\@let@token.\else.\null\fi\xspace}
\newcommand{\eg}{\emph{e.g}\onedot}
\newcommand{\ie}{\emph{i.e}\onedot}

\newcommand{\blueub}{}
\def\blueub#1{%
  \@ifnextchar_{\@blueub{#1}}{\@blueub{#1}_{}}}
\def\@blueub#1_#2{%
  \colorlet{currentcolor}{.}%
  \color{blue}%
  \underbrace{\color{currentcolor}#1}_{\color{blue}#2}%
  \color{currentcolor}}
\makeatother

\title{Aligning Thoughts with Answers: Probability Rewards to Tame Thinking Drift}
\author{%
\begin{minipage}[t]{0.96\textwidth}
\centering
\textbf{Pengzhan Sun$^{1,\dagger}$, Shiu-hong Kao$^{1,\dagger}$, Shijie Li$^{2}$, Yongyi Su$^{3}$,}\\
\textbf{Junbin Xiao$^{4,\ast}$, Arjun Reddy Akula$^{5}$, Angela Yao$^{1}$}\\[0.5em]
\normalfont\small
$^{1}$National University of Singapore  \quad
$^{3}$South China University of Technology \\
$^{2}$A*STAR Institute of Advanced Intelligence and Computing, Singapore\\
$^{4}$University of Science and Technology of China \quad
$^{5}$Google DeepMind
\\[0.4em]
{\scriptsize\ttfamily
\{pengzhan,ayao\}@comp.nus.edu.sg, shkao@u.nus.edu\\
junbinxiao@ustc.edu.cn, arjunakula@google.com\\
}
\end{minipage}%
}
\ifdefined\ritaArxivVersion
  \iclrfinalcopy
  \hypersetup{%
    pdfauthor={Pengzhan Sun, Shiu-hong Kao, Shijie Li, Yongyi Su, Junbin Xiao, Arjun Reddy Akula, Angela Yao},
    pdftitle={Aligning Thoughts with Answers: Probability Rewards to Tame Thinking Drift}%
  }
\fi
\begin{document}
\maketitle
\ifdefined\ritaArxivVersion
  \lhead{Preprint}
  \begingroup
    \renewcommand{\thefootnote}{\fnsymbol{footnote}}
    \footnotetext[2]{Equal contribution}
    \footnotetext[1]{Corresponding author}
  \endgroup
  \vspace{-0.2in}
\fi
\suppressfloats[t]

\begin{abstract}

This paper studies \textbf{thinking--answer consistency} in vision-language models.  We focus on Visual Intention Grounding, where a model infers a target object based on a human intention query and predicts a bounding box.
We reveal that previous IoU-based reinforcement learning (RL) frameworks suffer from ``thinking drift'', where the model produces a correct bounding box, despite having an incorrect reasoning process pointing to a different target object. Thus, we propose \textbf{Rita} (\textit{ReInforcing Thinking--Answer consistency}) as a novel RL paradigm to tame the drift. Specifically, Rita introduces two reasoning-label-free RL rewards, constructed from the conditional probability of reference answers: a \textbf{thinking reward} and a \textbf{consistency reward}. It also adopts a difficulty-aware \textbf{data filtering} strategy that selects informative easy-to-medium samples for RL using rollout error rate and reward variance. Extensive experiments on EgoIntention and the new RefEgo-Int benchmarks show that Rita performs consistently superior to the supervised finetuning approaches and vanilla RL-finetuned frameworks.

\end{abstract}

\section{Introduction}


\begin{wrapfigure}{R}{0.5\textwidth}
    \centering
    \includegraphics[width=\linewidth]{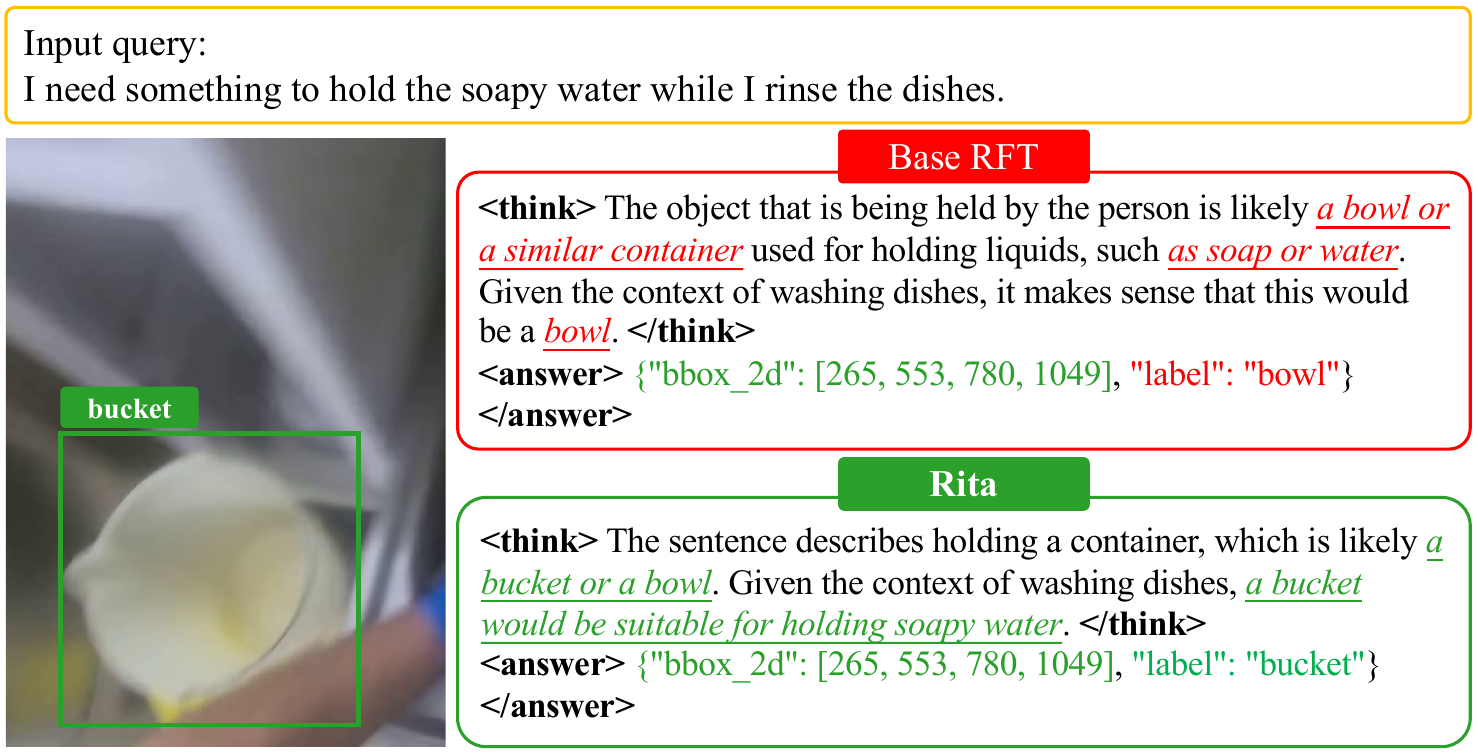}
    \caption{
        \textbf{Base RFT with IoU reward alone fails to enforce semantic consistency.}
        In the \textcolor{red}{Base RFT} (\textcolor{red}{red}), which relies solely on the IoU reward, the model predicts a reasonable bounding box but provides irrelevant reasoning, leading to a label error.
        In contrast, in our \textcolor{ForestGreen}{Rita} (\textcolor{ForestGreen}{green}), the integration of thinking and consistency rewards encourages consistency between the reasoning process and the final prediction, correcting the label while maintaining spatial accuracy.
    }
    \label{fig:teaser}
    \vspace{-0.15in}
\end{wrapfigure}

Given an egocentric image, Visual Intention Grounding (VIG)~\citep{Sun_2025_ICCV} requires a vision-language model (VLM) to identify and localize an object that can satisfy a user's intention.
Unlike an explicit referring expression, an intention query may describe a goal without naming the target object.
For example, a request for something to hold soapy water while rinsing dishes can refer to the bucket shown in Fig.~\ref{fig:teaser}.
Resolving such a query requires connecting the intended use with the objects and their context in the image.
This capability allows visual assistants to help users find suitable objects from descriptions of their goals, without requiring explicit object names or locations.

Recent VLMs optimized via reinforcement learning (RL)~\citep{liu_visual-rft_2025,shen_vlm-r1_2025} adopt a think-then-answer mechanism, generating a reasoning trace before producing the final prediction.
For VIG, this paradigm provides a natural decomposition: the model first interprets the user's intention and identifies a suitable target object in its reasoning trace, and then localizes that object in the final answer.

Existing RL-based grounding methods, such as VLM-R1~\citep{shen_vlm-r1_2025} and UniVG-R1~\citep{bai_univg-r1_2025}, optimize the Intersection over Union (IoU) between predicted and ground-truth bounding boxes.
Apart from an auxiliary reward for valid output formatting, their RL feedback evaluates only the final bounding box, without explicitly assessing consistency between the reasoning trace and the predicted target. Because the IoU reward depends only on the final bounding box, two responses receive the same reward when they predict the same box, even if their reasoning traces identify different target objects. This creates a risk of reward hacking, where the model can obtain a high reward through an inconsistent reasoning process, effectively predicting the correct box for the wrong reason.
For example, in Fig.~\ref{fig:teaser}, with an IoU-only objective, the target object is correctly located, but the \texttt{<think>} trace identifies the wrong object category.
We refer to this inconsistency as ``thinking drift'', inspired by~\cite{luo2025thinking}.
Worse still, once such reward hacking occurs, RL may reinforce the entire high-reward rollout, thereby exacerbating the inconsistency. Indeed, when we train UniVG-R1 on VIG, IoU-only RL improves grounding precision, yet 15.8\% of the boxes it newly corrects come with reasoning that targets a different object, $2.4\times$ the drift rate before RL (Appendix~\ref{sec:drift_analysis}).
One remedy is to supervise the reasoning process, \eg, using human-labeled rationales~\citep{shao2024visual} or chains of thought (CoT)~\citep{wei2022chain} distilled from stronger teacher models~\citep{guo2025deepseek,achiam2023gpt}, as in the cold-start stage of UniVG-R1~\citep{bai_univg-r1_2025}.
However, such supervision either requires costly annotation at scale~\citep{zhao2025unsupervised} or risks overfitting the model to the limited reasoning patterns of the teacher models~\citep{chen2023mcc,matos2025cognitive}.


To tackle thinking drift, we propose a reasoning label-free RL paradigm \textbf{Rita} (\textit{ReInforcing Thinking--Answer consistency}).
Motivated by \citet{welch2026cost}, who show that a VLM's post-reasoning answer likelihood increasingly reflects its reasoning-answer consistency rather than calibrated confidence in answer correctness, we turn this consistency signal into a correctness signal by scoring the reference answer instead of the sampled one. Specifically, our \textbf{thinking reward} encourages reasoning traces which directly contribute to higher conditional likelihood to the reference answer.
Additionally, because this substitution detaches the thinking reward from the generated answer, simply adding it to the IoU reward does not ensure that a reference-supporting trace and a correct prediction occur within the same rollout; a high score from one reward may compensate for a low score from the other. We therefore introduce a \textbf{consistency regularization} that explicitly couples trace and answer quality, rewarding a reference-supporting trace when the rollout's predicted box overlaps the target and penalizing it when the box largely misses the target.

Beyond improving reasoning reliability, we investigate strategic data filtering to enhance training efficiency.
Some reasoning-focused RL methods prioritize challenging but solvable problems to elicit deeper reasoning~\citep{wang_sota_2025}.
For grounding, however, we find it more effective to concentrate on samples of low-to-moderate difficulty.
Medium-difficulty samples show the greatest variation in rewards, while samples with no successful rollout provide little contrast on average for policy updates.
By filtering toward samples with informative reward variation, our strategy improves grounding performance with RL fine-tuning on 5\% of the original SFT training set.

We evaluate Rita on two VIG benchmarks: EgoIntention~\citep{Sun_2025_ICCV} and RefEgo-Int, a new dataset curated from RefEgo~\citep{kurita_refego_2023}.
On Qwen2.5-VL-3B / 7B-Instruct~\citep{bai2025qwen2}, Rita yields consistent gains in both reasoning and grounding accuracy.
Our contributions are as follows:
\begin{itemize}
  \item We identify \emph{thinking drift} in IoU-rewarded RL for visual intention grounding: a rollout can localize the target while its reasoning trace selects a different object, and outcome-only rewards cannot distinguish it from a consistent rollout.
  \item We propose Rita, which adds two reasoning-label-free rewards computed from the policy's likelihood of the reference answer given a sampled trace: a thinking reward that scores how strongly the trace supports the reference answer, and a consistency reward that couples this score with the IoU of the rollout's own prediction.
  \item We introduce a rollout-based data filtering strategy that trains on easy-to-medium samples with high reward variance, using 5\% of the SFT training set.
  \item Rita improves P@0.5 over SFT Qwen2.5-VL by +4.8 (3B) and +2.1 (7B) on EgoIntention and by +10.2 and +7.5 zero-shot on RefEgo-Int, raises grounded reasoning accuracy over IoU-only RL, and outperforms UniVG-R1 in P@0.5 without distilled reasoning traces.
\end{itemize}





\vspace{-5pt}
\section{Related Work}
\vspace{5pt}\noindent{\bf RL for Vision Language Models.}
Recent industrial foundation models such as DeepSeek-R1~\citep{guo2025deepseek}, Kimi k1.5~\citep{team2025kimi}, and OpenAI's o1 system~\citep{jaech2024openai} have popularized GRPO-style reinforcement learning (RL) as a scalable post-training recipe, establishing the think-then-answer paradigm and triggering a wave of RL research on vision--language models (VLMs). Most subsequent academic efforts apply RL to tasks that naturally require long-chain reasoning, including mathematical and scientific QA, visual reasoning, and spatial QA, where multi-step CoT-style solutions are essential~\citep{wang2025perception,jiang_vlm-r3_2025,meng_mm-eureka_2025,sarch2025grounded,tan_reason-rft_2025}. Closer to our work, recent perception-centric works (\eg, recognition, OCR, grounding, segmentation) show that think-then-answer VLMs can benefit from RL when the final objective is a perceptual metric~\citep{yu_perception-r1_2025,ma_one_2025,shen_vlm-r1_2025,bai_univg-r1_2025,liu2025seg}.

These methods reward only the final prediction and leave the content of the reasoning trace unconstrained.
Methods that supervise the trace require additional reasoning data, such as human-annotated rationales~\citep{shao2024visual} or chains of thought distilled from stronger models, as in the cold-start stage of UniVG-R1~\citep{bai_univg-r1_2025}.
Such data are costly to collect at scale~\citep{zhao2025unsupervised}, and distilled traces can inherit the limited reasoning patterns of the teacher~\citep{chen2023mcc,matos2025cognitive}.
Rita instead rewards the trace using only the existing answer annotations.

\vspace{5pt}\noindent{\bf Label-free RL Rewards.}
To extend RL beyond tasks with rule-based verifiers, recent work trains language models on unverifiable data~\citep{wang_native_2026,tang_beyond_2025}.
Reference-likelihood rewards replace the verifier with the policy's own likelihood of the reference answer after its reasoning, and use this signal to learn correct answers in question answering~\citep{zhou_reinforcing_2025,yu_rlpr_2025,liu_nover_nodate}.
In grounding, IoU already verifies the correctness of the answer, so Rita uses reference likelihood for a different purpose.
Because answer likelihood after reasoning reflects consistency with the trace~\citep{welch2026cost}, it measures whether the reasoning trace supports the correct target.
Rita further adds a consistency reward that ties this trace score to the model's own prediction, which reference-likelihood rewards discard.

\vspace{5pt}\noindent{\bf Data Filtering in RL.}
Early instruction-tuning studies already demonstrate that carefully filtered, task-specialized subsets can outperform using the full corpus \citep{xia_less_2024, liu_what_2024, li_quantity_2024, li_superfiltering_2024, jain_llm-assisted_2023, chen_maybe_2023, zhou2023lima}. Particularly, recent work, such as Vision-G1~\citep{zha_vision-g1_2025}, QwenLong-L1~\citep{wan_qwenlong-l1_2025}, and ThinkLite-VL~\citep{wang_sota_2025}, adopt various difficulty-aware or influence-based criteria to prioritize ``hard-but-solvable'' or unsolved examples, under the assumption that such instances provide the strongest learning signal for reasoning tasks (\eg, math, logic, long-context QA). Different from these approaches, we propose a selection scheme that favors informative samples of easy-to-medium difficulty, avoiding the noisy, uninformative gradients from persistent failures like near-zero IoU rewards and thereby performing a more stable optimization curriculum.


\section{Method}




\subsection{Preliminary: RL for Visual Intention Grounding}
\label{3.1}

\noindent{\bf Visual Intention Grounding} maps an intention sentence to the target object and its location. Given an input image $I$ and a human intention query $Q$, the model infers the underlying intention and predicts an answer $A$ that contains the bounding box $(x_1,y_1,x_2,y_2)$ of the target object. Recent RL-finetuned reasoning models first generate a thinking trace $T$, \ie, $(T,A)\sim\pi_\theta(\cdot\mid I,Q)$, where $T$ describes how the model links the query $Q$ to the predicted answer $A$.

\begin{figure}[t] 
    \centering
    \includegraphics[width=\linewidth]{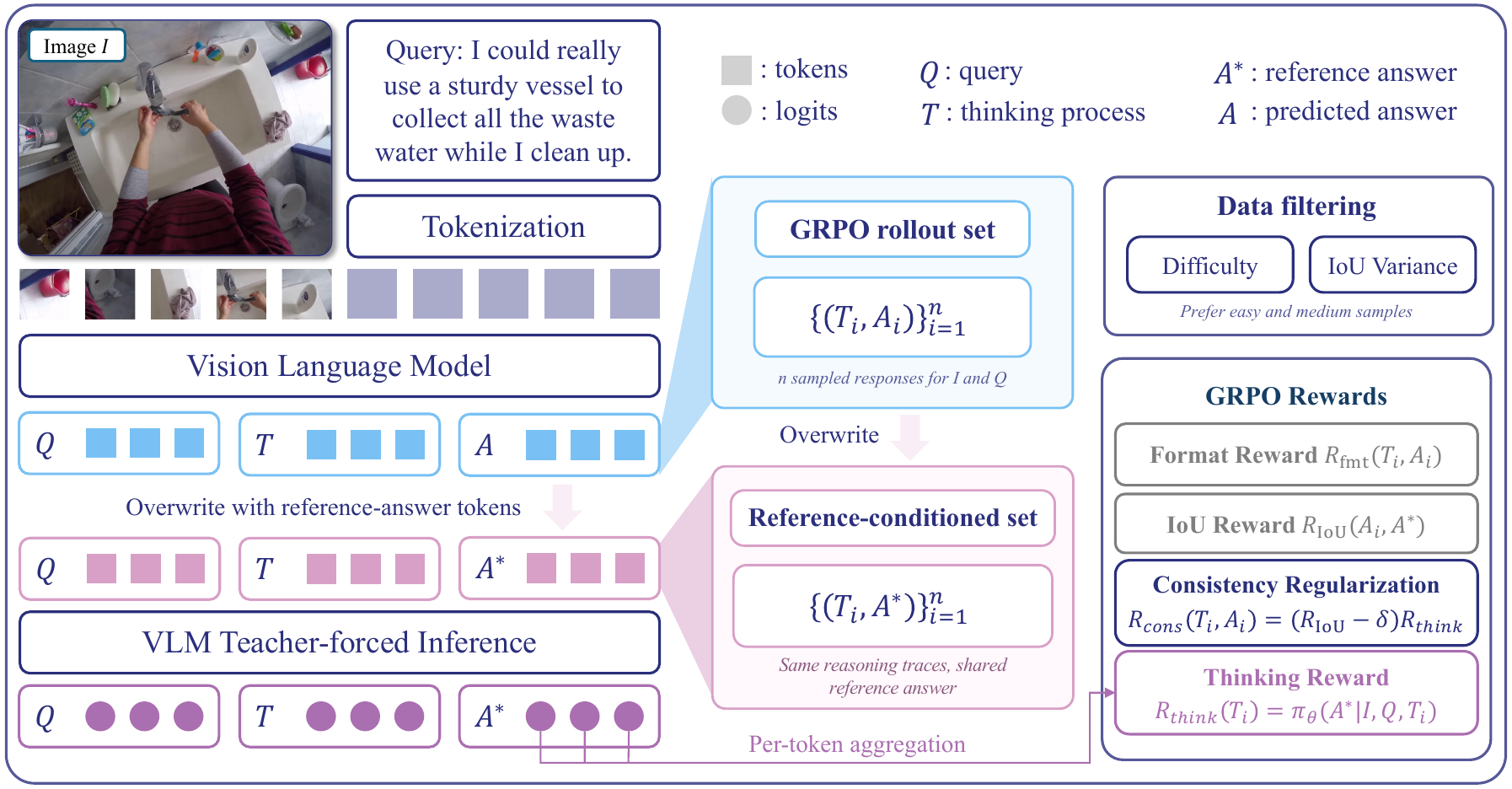}
    \caption{
    \textbf{Overview of the Rita framework.} Given an input image and human intention query, the VLM samples multiple think-then-answer rollouts $(Q, T_k, A_k)$, from which we derive a grounding reward via IoU with the ground-truth box and a format reward. We then overwrite $A_k$ with the reference answer $A^\star$ and run the VLM in teacher-forced mode to obtain logits for the answer tokens, which define the thinking reward $R_{\mathrm{think}}$. The consistency reward $R_{\mathrm{cons}}$ scales this score by the predicted box's IoU minus a margin $\delta$. During the GRPO-based optimization, we conduct difficulty-and-variance-aware data sampling computed from IoU rollouts to improve training efficiency.
    }
    \label{fig:thinking_reward}

\end{figure}

\vspace{5pt}\noindent{\bf GRPO for VLMs.}
Group Relative Policy Optimization (GRPO) is a rule-based RL algorithm for post-training LLMs/VLMs that removes the need for a learned critic by using group-relative rewards. 
For each input \(q\), the old policy samples a group of \(N\) candidate outputs \(\{o_i\}_{i=1}^N\) and each reward $r_i$ is normalized within the group into the relative advantage $A_i=(r_i-\mu)/\sigma$.
The GRPO objective is
\begin{equation}
\mathcal{J}_{\mathrm{GRPO}}(\theta) = \mathbb{E}_{\{o_i\}\sim \pi_{\theta_{\text{old}}}(q)} \Bigg[
\frac{1}{N}\sum_{i=1}^N
\Big\{
\min\!\big[s_1 A_i,\; s_2 A_i\big]
 -  \beta\,\mathbb{D}_{\mathrm{KL}}\!\big[\pi_\theta \,\|\, \pi_{\mathrm{ref}}\big]
\Big\}
\Bigg],
\end{equation}
where $s_1=\frac{\pi_\theta(o_i\mid q)}{\pi_{\theta_{\text{old}}}(o_i\mid q)}$ and $s_2  =\operatorname{clip}\!\left(\frac{\pi_\theta(o_i\mid q)}{\pi_{\theta_{\text{old}}}(o_i\mid q)},\, 1-\epsilon,\, 1+\epsilon\right)$ and $\beta$ weights the KL penalty toward the reference policy $\pi_{\mathrm{ref}}$.

\vspace{5pt}\noindent{\bf RL Rewards.}
Early RL approaches~\citep{liu_visual-rft_2025,shen_vlm-r1_2025,bai_univg-r1_2025} for VLM perception tasks optimize verifiable rewards aligned with the evaluation metric:
\begin{equation}\label{eq:reward_vanilla}
    R(T, A\mid A^\star,\lambda_1,\lambda_2)\;=\;\lambda_1 R_\text{IoU}(A, A^\star)+\lambda_2 R_\text{fmt}(T,A).
\end{equation}
Here, $R_{\text{IoU}}\in[0,1]$ is the IoU of two bounding boxes, and $R_{\text{fmt}}\in\{0,1\}$ is a binary indicator of whether $(T,A)$ follows the required output format, \ie, \texttt{<think>}\ldots{} \texttt{</think>} \texttt{<answer>}\ldots{}\texttt{</answer>}. 
This answer-oriented reward can produce inconsistent thinking traces and answers: the reasoning content is learned only indirectly through the IoU reward, and neither the trace nor its consistency with the answer is supervised.
In our re-implementation of UniVG-R1, IoU-only RL indeed increases the share of correct boxes paired with drifted reasoning (Appendix~\ref{sec:drift_analysis}).
To address this without reasoning labels, we directly optimize the thinking trace and thought--answer consistency with the probability-based rewards in Section~\ref{3.2}.

\subsection{Rita: Outcome-Conditioned Probability-based Rewards}
\label{3.2}

Standard grounding rewards evaluate only the final prediction. Consequently, two rollouts receive the same IoU reward whenever they predict the same bounding box, even if one reasoning trace correctly identifies the intended object while the other refers to an unrelated target. To provide direct feedback on the generated trace without requiring reasoning annotations, Rita evaluates whether the trace supports the annotated answer. 

\paragraph{Thinking reward.}
Let \(x=(I,Q)\) denote an image--query pair, and let a rollout from the old policy be \(o_k=(T_k,A_k)\), where \(T_k\) is the generated thinking trace and \(A_k\) is the sampled structured answer. Given the reference answer
\(A^\star=(a_1^\star,\ldots,a_L^\star)\), we preserve the sampled trace \(T_k\), replace \(A_k\) with \(A^\star\), and perform a teacher-forced forward pass, as illustrated in Fig.~\ref{fig:thinking_reward}. We compute the length-normalized likelihood score
\begin{equation}
R_{\mathrm{think}}
=
\frac{1}{L}
\sum_{t=1}^{L}
\pi_{\theta}
\left(
a_t^\star
\mid x,T_k,a_{<t}^\star
\right),
\label{eq:answer_loglikelihood}
\end{equation}
where \(\pi_{\theta}\) is the policy snapshot used for reward computation. Intuitively, Eq.~\ref{eq:answer_loglikelihood} averages the token-level probabilities assigned to the reference answer, conditioned on the given input $(I,Q)$ and sampled thinking trace $T_k$.

Overwriting with the reference answer is important because, after a reasoning trace, the likelihood of the sampled answer \(A_k\) mainly reflects its consistency with the trace rather than its correctness~\citep{welch2026cost}, so it would reward confident incorrect predictions as well (see Appendix~\ref{sec:overwrite_ablation} for an ablation). In contrast, Eq.~\ref{eq:answer_loglikelihood} holds the target answer fixed across all rollouts of the same input. Differences in reward therefore reflect how strongly each sampled trace supports the annotated target. \(R_{\mathrm{think}}\) thus provides a reasoning-label-free signal for comparing traces.

\paragraph{Thinking-answer consistency regularization.}
Although \(R_{\mathrm{think}}\) evaluates whether \(T_k\) supports \(A^\star\), it does not consider the answer \(A_k\) actually generated after that trace. Conversely, \(R_{\mathrm{IoU}}\) evaluates the sampled answer without considering whether its preceding trace provides compatible reasoning. Simply adding these rewards therefore promotes trace quality and answer quality as two separate objectives, but does not explicitly reward their joint satisfaction within the same rollout. Consequently, a high score from one component can compensate for a low score from the other, so the sum does not address thinking drift.

To explicitly couple these two components, we introduce a thinking--answer consistency regularization. We define the box score
\begin{equation}
S_\delta(A_k,A^\star)
=R_\text{IoU}(A_k,A^\star)-\delta,
\label{eq:answer_quality}
\end{equation}
where \(\delta\) is an IoU margin that separates rollouts whose predicted box overlaps the target from rollouts whose box misses it.
We use a small margin, \(\delta=0.1\), so that the penalty applies only to boxes with little or no overlap with the target; Appendix~\ref{sec:delta_ablation} compares it with the standard grounding threshold \(\delta=0.5\). We define
\begin{equation}
R_{\mathrm{cons}}
=
S_\delta(A_k,A^\star)
R_{\mathrm{think}}.
\label{eq:consistency_reward}
\end{equation}
This multiplicative form couples trace quality and answer quality. When the predicted box overlaps the target by more than \(\delta\), \(S_\delta>0\), and a trace that strongly supports the reference answer receives a larger positive consistency reward. When the predicted box misses the target, \(S_\delta<0\), a high thinking score instead produces a stronger penalty, reflecting the inconsistency between the reference-supporting trace and the realized answer. If the trace itself provides little support for the reference answer, \(R_{\mathrm{cons}}\) remains small regardless of the answer quality. Therefore, the consistency reward becomes strongly positive only when both components agree, while providing little or negative reinforcement when an inconsistency is observed.

\paragraph{Overall objective.}
For rollout \(k\), Rita combines the proposed probability rewards with the standard grounding and formatting rewards:
\begin{equation}
\begin{split}
R
={}&
\lambda_{\mathrm{IoU}}R_{\mathrm{IoU}}
+
\lambda_{\mathrm{fmt}}R_{\mathrm{fmt}}
+
\lambda_{\mathrm{think}}R_{\mathrm{think}}
+
\lambda_{\mathrm{cons}}R_{\mathrm{cons}}.
\end{split}
\label{eq:rita_reward}
\end{equation}
The four terms provide complementary supervision: \(R_{\mathrm{IoU}}\) evaluates spatial accuracy, \(R_{\mathrm{fmt}}\) enforces valid output structure, \(R_{\mathrm{think}}\) evaluates whether the sampled trace supports the annotated target, and \(R_{\mathrm{cons}}\) couples this compatibility signal to the answer produced by the rollout. All rewards are computed without reasoning labels and normalized within each rollout group by GRPO.

\subsection{Rita: Data Filtering}
\label{3.3}


While our probability-based rewards provide richer feedback for individual rollouts, effective policy optimization also requires informative training samples. Rollout groups dominated by unsuccessful grounding predictions often provide limited outcome-level contrast. We therefore introduce a difficulty- and variance-aware data selection strategy that prioritizes learnable samples with diverse grounding outcomes, improving data efficiency while complementing our reward design.

For each training sample $i$, we sample $N$ rollouts and compute an IoU-based verifiable reward $r_{i}^{(k)}\in[0,1]$ for rollout $k$, with correctness indicator $c_{i}^{(k)}=\mathbb{I}\!\left[\text{IoU}_{i}^{(k)}\ge \tau\right]$ under a fixed threshold $\tau$. We define a sample difficulty score as the number of correct rollouts $d_i=\sum_{k=1}^{N} c_{i}^{(k)}$ (smaller $d_i$ means harder), and partition the dataset into three buckets: hard $B_h=\{i\,|\,d_i\in[0,\gamma_h N]\}$, medium $B_m=\{i\,|\,d_i\in(\gamma_{h} N,\gamma_{e}N)\}$, and easy $B_e=\{i\,|\,d_i\in[\gamma_e N,N]\}$, where $\gamma_h,\gamma_e$ are hyperparameters for bucket thresholds with $0\leq\gamma_h<\gamma_e\leq 1$.

In addition to difficulty, we quantify a sample's informativeness by the reward spread $\sigma_i=\operatorname{Std}\!\left(\{r_{i}^{(k)}\}_{k=1}^{N}\right)$, which equals the dispersion of the rollout advantages before normalization: for any baseline $b_i$, the unnormalized advantages $\tilde{A}_{i}^{(k)}\!=\!r_{i}^{(k)}\!-\!b_i$ satisfy $\operatorname{Std}(\{\tilde{A}_{i}^{(k)}\}_{k=1}^{N})=\sigma_i$. Intuitively, a larger $\sigma_i$ indicates a greater separation between good and bad rollouts, yielding stronger policy gradients for GRPO-style updates.

Our filtering-and-sampling scheme specifies a portion vector $\boldsymbol{\pi}=(\pi_h,\pi_m,\pi_e)$ with $\pi_h+\pi_m+\pi_e=1$, and draws $M\pi_b$ samples from bucket $B_b$, where $M$ is the size of the RL training set, according to a variance-biased distribution
\begin{equation}
P(i\mid B_b)=\frac{\sigma_i^{\alpha}}{\sum_{j\in B_b}\sigma_j^{\alpha}},\quad \alpha>0,
\end{equation}
where $\alpha$ is a temperature controlling within-bucket sharpness and $b\in\{h,m,e\}$.
To avoid misleading updates from unlearnable cases, we discard samples with $d_i=0$ regardless of high $\sigma_i$; this prevents the optimizer from chasing high-variance but consistently incorrect rollouts. In practice, $\boldsymbol{\pi}$ emphasizes $B_m$ for learnability, adds $B_h$ for exploration and $B_e$ for stability, and the $\sigma_i^{\alpha}$ weighting favors samples with larger advantage gaps.

\section{Experiments}
\label{sec:experiments}

\subsection{Implementation Details}

\textbf{Datasets.}
Our evaluation focuses on egocentric VIG datasets derived from head-mounted-camera imagery~\citep{grauman_ego4d_2022}. Such imagery presents characteristic challenges, including frequent camera motion, motion blur, and off-center target objects. Our primary benchmark is
EgoIntention~\citep{Sun_2025_ICCV}, a visual grounding benchmark pairing Ego4D images with intention sentences. It contains two splits: \textit{context} (typical intents, \eg, ``I need to sit down'' $\rightarrow$ chair) and \textit{uncommon} (atypical intents, \eg, ``I need a boost to change the bulb'' $\rightarrow$ chair).
To assess generalization, we construct RefEgo-Int from RefEgo~\citep{kurita_refego_2023} by sampling frames (one frame per video clip) and collecting context intention sentences following the EgoIntention dataset protocol. RefEgo-Int preserves RefEgo key challenges: (i) frequent viewpoint motion, (ii) no-target cases, and (iii) long-tail object categories. Refer to Appendix~\ref{sec:refego} for RefEgo-Int construction details.

\vspace{5pt}\noindent \textbf{Models.}
We use Qwen2.5-VL-3B-Instruct as the main backbone and report scalability with Qwen2.5-VL-7B-Instruct~\citep{bai2025qwen2}.
We follow VLM-R1~\citep{shen_vlm-r1_2025} for the GRPO settings, with $N{=}8$ rollouts, temperature $0.9$, iterations $1$, KL ratio $0.04$, and learning rate $1\mathrm{e}{-6}$. All reward weights $\lambda_i$'s are set to $1$, and the IoU margin of the consistency reward is $\delta=0.1$. For data filtering, we set $\tau=0.5$, $\gamma_h=0.2$ and $\gamma_e=0.8$.
The grounding template and thinking prompts also follow VLM-R1 (Appendix~\ref{sec:prompt}).

For the SFT baseline, we perform full finetuning with LLaMA-Factory~\citep{zheng2024llamafactory} on the EgoIntention training set for 1 epoch, whereas for RL finetuning we apply Low-Rank Adaptation (LoRA)~\citep{hu2022lora} (rank 64) for 500 update steps.
We re-implement UniVG-R1~\citep{bai_univg-r1_2025} on EgoIntention with the same Qwen2.5-VL-3B-Instruct backbone.
Following its two-stage recipe, we first fine-tune the backbone for one epoch on chain-of-thought traces distilled from Qwen2.5-VL-72B-Instruct~\citep{bai2025qwen2} (replacing the proprietary Qwen-VL-Max teacher), keeping only traces whose final answers are correct.
We then run its GRPO stage with the official implementation, including its IoU and format rewards and difficulty-aware weight adjustment, under the same RL configuration as Rita.


\vspace{5pt}\noindent \textbf{Metrics.} We evaluate our models with three metrics:
\textit{Grounding Precision} (P@0.5) measures spatial accuracy, where a predicted box is correct if its IoU with the ground truth is $\ge 0.5$.
\textit{Mean Intersection over Union} (mIoU) provides a fine-grained measure of localization quality by averaging the IoU
across all samples.
\textit{Thinking Accuracy} (Acc$_{\text{think}}$) assesses the logical correctness of the thinking process by checking if the predicted object category matches the ground-truth label.
\textit{Grounded Reasoning Accuracy} (Acc$_{\text{align}}$) is a holistic metric that requires a sample to have both a correct bounding box ($IoU \ge 0.5$) \textit{and} the correct object category.  ~\AY{given that your main motivation was to reduce the number of instances of misaligned thinking processes, this needs to be measured with and without your proposed reward to show how it had an effect no?  otherwise, how can we really say that your gains come from that?}

\subsection{Quantitative Results}

\begin{table}[t]
  \centering
  \caption{\textbf{EgoIntention results split by Context / Uncommon and Overall.}
  Metrics are \textit{P@0.5} (\ensuremath{\uparrow}) and \textit{mIoU} (\ensuremath{\uparrow}). Numbers in parentheses give the P@0.5 gain of Rita over the Qwen2.5-VL-Instruct baseline of the same size.  ~\AY{I'm surpised by the improvement over our previous work; do they really come from this type of strategy or is it simply the RL?  again, i think this table should also have the same evaluation measurements as Table 2, to convince the reader that your grounding is good, but thinking trace is also correct, since that is your main motivation?}}
  \label{tab:egointention_main}
  \centering
 \resizebox{\linewidth}{!}{%
  \begin{tabular}{
    l
    S[table-format=2.1] S[table-format=1.3]
    S[table-format=2.1] S[table-format=1.3]
    S[table-format=2.1] S[table-format=1.3]
  }
    \toprule
    \multirow{2}{*}{Method}
      & \multicolumn{2}{c}{\textbf{Context}}
      & \multicolumn{2}{c}{\textbf{Uncommon}}
      & \multicolumn{2}{c}{\textbf{Overall}} \\
    \cmidrule(lr){2-3} \cmidrule(lr){4-5} \cmidrule(lr){6-7}
      & {P@0.5} & {mIoU} & {P@0.5} & {mIoU} & {P@0.5} & {mIoU} \\
    \midrule
    Qwen-VL~\citep{bai_qwen-vl_2023}                & 32.1 & 0.350 & 26.1 & 0.298 & 29.1 & 0.324 \\
    MiniGPT-v2~\citep{chen_minigpt-v2_2023}         & 46.0 & 0.429 & 40.9 & 0.373 & 43.4 & 0.401 \\
    Reason-to-Ground~\citep{Sun_2025_ICCV}          & 49.9 & 0.445 & 44.7 & 0.396 & 47.3 & 0.421 \\
    Qwen2.5-VL-3B-Instruct~\citep{bai2025qwen2}     & 64.8 & 0.581 & 56.1 & 0.483 & 60.4 & 0.532 \\
    Qwen2.5-VL-7B-Instruct~\citep{bai2025qwen2}     & 67.2 & 0.594 & 61.8 & 0.514 & 64.5 & 0.554 \\
    UniVG-R1 (Qwen2.5-VL-3B)~\citep{bai_univg-r1_2025}&65.9& 0.604 & 59.6 & 0.549 & 62.7 & 0.576 \\
    \midrule
    \rowcolor{gray!8}
    Rita-3B & {67.9\,{\scriptsize\textcolor{ForestGreen}{(+3.1)}}} & 0.598 & {62.4\,{\scriptsize\textcolor{ForestGreen}{(+6.3)}}} & 0.518 & {65.2\,{\scriptsize\textcolor{ForestGreen}{(+4.8)}}} & 0.558 \\
    \rowcolor{gray!8}
    Rita-7B & {68.9\,{\scriptsize\textcolor{ForestGreen}{(+1.7)}}} & 0.605 & {64.3\,{\scriptsize\textcolor{ForestGreen}{(+2.5)}}} & 0.527 & {66.6\,{\scriptsize\textcolor{ForestGreen}{(+2.1)}}} & 0.566 \\
    \bottomrule
  \end{tabular}%
  }

\end{table}


\vspace{5pt}\noindent\textbf{Localization Evaluation.}
We evaluated our method on the EgoIntention \textit{Context} split and \textit{Uncommon} split, reporting overall accuracy as shown in Table~\ref{tab:egointention_main}.
Prior methods, Qwen-VL~\citep{bai_qwen-vl_2023}, MiniGPT-v2~\citep{chen_minigpt-v2_2023}, and Reason-to-Ground~\citep{Sun_2025_ICCV}, which disentangles intention reasoning from localization, achieve limited performance.
In comparison, our base model Qwen2.5-VL-3B-Instruct~\mbox{\citep{bai2025qwen2}} achieves substantial improvements.
Building on the supervised finetuning models, Rita-3B improves precision@0.5 by 3.1 on the context split and by 6.3 on the uncommon split.
For 7B models, the context split increases by 1.7 and the uncommon split by 2.5.
These results demonstrate that Rita is effective even when compared to strong supervised fine-tuning baselines at both 3B and 7B scales. Relative to these baselines, mIoU follows the same trend as precision@0.5.
Rita-3B also outperforms UniVG-R1~\citep{bai_univg-r1_2025} on the same backbone by 2.0/2.8 precision@0.5 on the context/uncommon splits without teacher-distilled reasoning traces, although UniVG-R1 attains a higher mIoU.

\begin{table}[t]
\centering
\caption{
    \textbf{Effectiveness of reward components on reasoning performance.}
    We evaluate Acc$_{think}$ and Acc$_{align}$ (\%) across \textit{Context} and \textit{Uncommon} splits. $R_{\text{IoU}}$, $R_{\text{think}}$, and $R_{\text{cons}}$ represent grounding, thinking, and consistency rewards used during RFT. Best results are in \textbf{bold}.
}

\label{tab:ablation_rewards_reasoning}
\centering
\begin{tabular}{@{}ccc|cc|cc@{}}
\toprule
\multicolumn{3}{c|}{\textbf{RFT Rewards}} & \multicolumn{2}{c|}{\textbf{Context Split (\%)}} & \multicolumn{2}{c}{\textbf{Uncommon Split (\%)}} \\
\cmidrule(lr){1-3} \cmidrule(lr){4-5} \cmidrule(l){6-7}
$R_{\text{IoU}}$ & $R_{\text{think}}$ & $R_{\text{cons}}$ & Acc$_\text{think}$ & Acc$_\text{align}$ & Acc$_\text{think}$ & Acc$_\text{align}$ \\ \midrule
\checkmark &            &            & 83.7    & 61.4    & 71.5    & 54.4     \\
\checkmark &            & \checkmark & 83.1    & 61.2    & 70.9    & 54.0    \\
\checkmark & \checkmark &            & 85.1    & 62.3    & 73.7    & 55.6    \\
\rowcolor[HTML]{EFEFEF}
\checkmark & \checkmark & \checkmark & \textbf{86.6} & \textbf{63.7} & \textbf{75.0} & \textbf{56.5} \\
\bottomrule
\end{tabular}%

\end{table}

\vspace{5pt}\noindent\textbf{Reasoning evaluation.}
Table~\ref{tab:ablation_rewards_reasoning} ablates the reward components used during RFT and reports reasoning quality (Acc$_{\text{think}}$) and thought--answer consistency (Acc$_{\text{align}}$) on both \textit{Context} and \textit{Uncommon} splits.
Adding the thinking reward $R_{\text{think}}$ yields a clear and consistent improvement over the $R_{\text{IoU}}$-only baseline: Acc$_{\text{think}}$ increases from 83.7\% to 85.1\% on \textit{Context} and from 71.5\% to 73.7\% on \textit{Uncommon}, while Acc$_{\text{align}}$ improves from 61.4\% to 62.3\% and from 54.4\% to 55.6\%, respectively.
In contrast, using the consistency reward $R_{\text{cons}}$ alone does not help and slightly degrades performance (\eg, 83.7\%$\rightarrow$83.1\% on Acc$_{\text{think}}$ for \textit{Context}).
We attribute this to the fact that $R_{\text{cons}}$ is defined relative to the direction induced by $R_{\text{think}}$; without $R_{\text{think}}$ participating in GRPO updates, the consistency signal cannot effectively steer the model toward better reasoning.
In other words, enforcing consistency between reasoning traces and answers, without learning informative reasoning, may instead hinder optimization and degrade the answer performance.
Finally, combining $R_{\text{cons}}$ with $R_{\text{think}}$ delivers the best results across both splits, improving Acc$_{\text{think}}$ to 86.6\%/75.0\% and Acc$_{\text{align}}$ to 63.7\%/56.5\% on \textit{Context}/\textit{Uncommon}.
Relative to $R_{\text{think}}$ alone, adding $R_{\text{cons}}$ raises Acc$_{\text{think}}$ by 1.5/1.3 and Acc$_{\text{align}}$ by 1.4/0.9 points, showing that the consistency reward complements the thinking reward even though it does not help on its own.

\subsection{Data Filtering}

We broadly explore multiple data sampling methods for Rita on visual intention grounding tasks.
\textit{Random.} The baseline randomly samples the RL training set.
\textit{Unsolved.} Following prior RL-for-reasoning works, we keep only samples that the SFT model fails on, these form the hard unsolved RL set.
\textit{Near threshold} (IoU $\approx 0.5$)\textit{.} We select samples whose SFT IoU is near the 0.5 decision boundary, assuming they are most correctable by RL.
\textit{Normal distribution sampling.} Using our difficulty (wrong rollout count) and informativeness (reward variance) estimates, we sample with a normal distribution centered at 4 wrong rollouts (midpoint of 0--8).
\textit{Rita Strategy.} Motivated by the drop with hard--unsolved data, we downweight hard cases and exclude all-wrong samples, focusing RL on informative easy--medium instances.

\Needspace{22\baselineskip}
\begin{wrapfigure}{r}{0.5\textwidth}
    \centering
    \includegraphics[width=\linewidth]{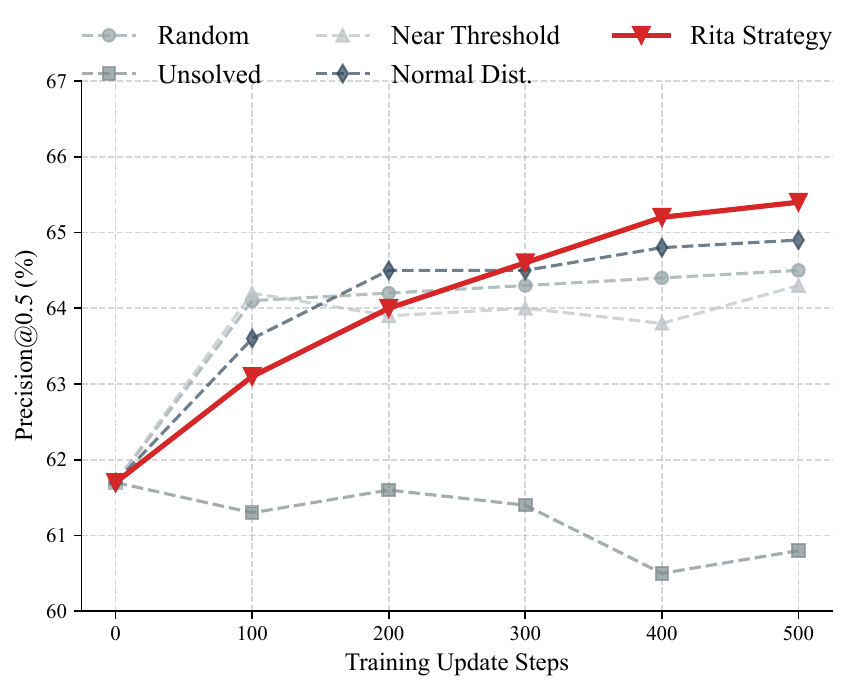}
    \caption{Grounding accuracy (Precision@0.5) on the EgoIntention dataset for five data filtering strategies across training update steps.}
    \label{fig:data_filtering}
\end{wrapfigure}

As shown in Fig.~\ref{fig:data_filtering}, we first observe that directly borrowing strategies from previous RL work on reasoning tasks does not work well. Using unsolved hard samples for RL finetuning cannot even boost the final grounding precision, performance drops from 61.7 at step 0 to 60.8 at step 500.
Compared with the baseline method without any data filtering strategy, we see a clear gap at step 500, dropping from 64.5 to 61.7.
For training samples with IoU near 0.5, there is a consistent slight decrease across updating steps compared with baseline results.
For the normal distribution approach based on the difficulty level we estimated for the training set, our strategy prefers samples with higher informativeness levels within each difficulty bin. We observe better results compared with the baseline method. At step 500, the overall grounding performance on EgoIntention outperforms the baseline method by 0.9.
Finally, we exclude samples with all wrong rollouts and use a sample ratio of 1:6:3 for hard, medium, and easy level samples. This achieves the best overall performance compared with baseline.

\subsection{Qualitative Results}

\begin{figure}[t] 
    \centering
    \includegraphics[width=0.9\linewidth]{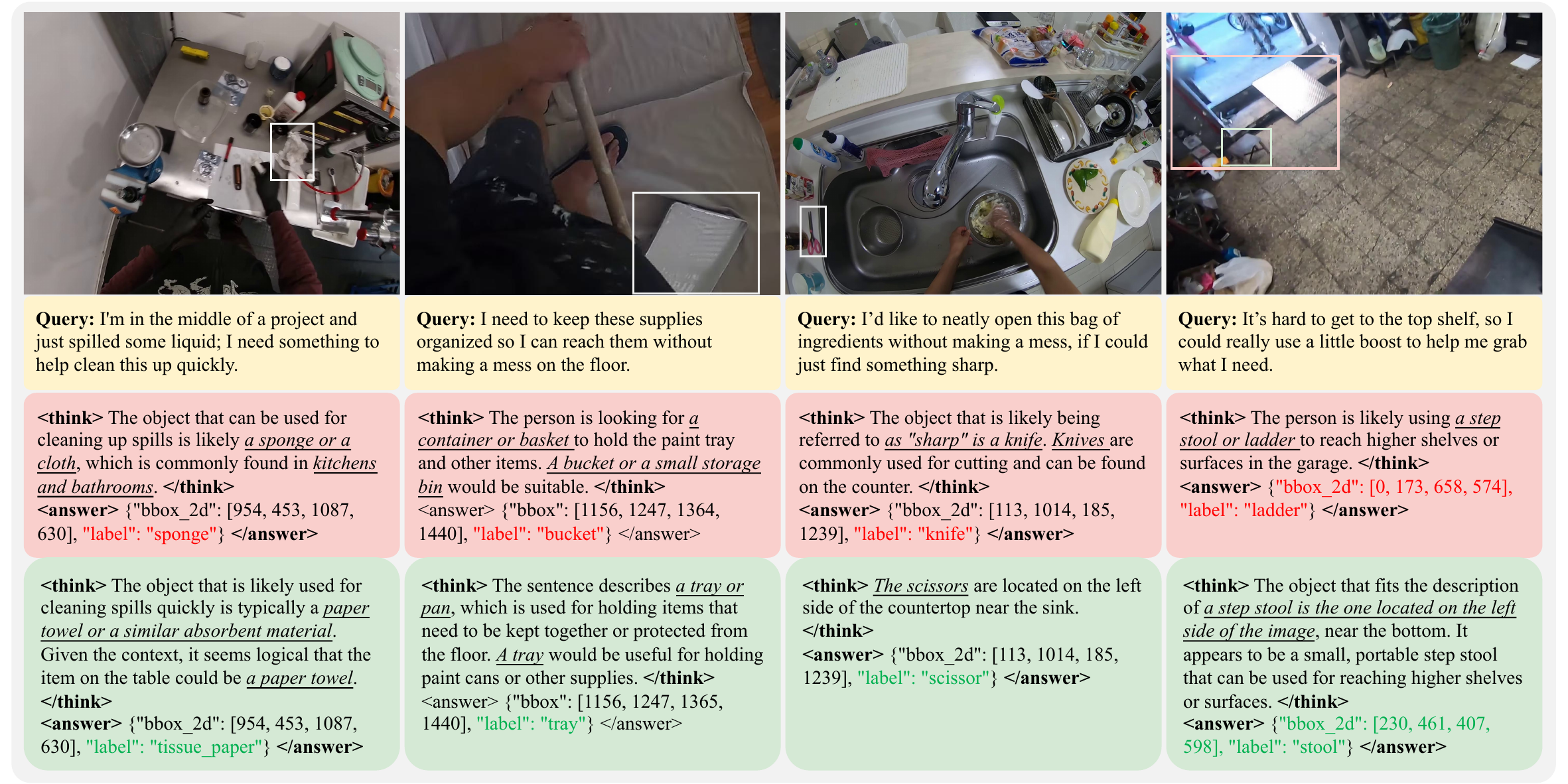}
    \caption{
        Qualitative comparison of the IoU-reward baseline (pink) and Rita (green), with key reasoning phrases \underline{underlined}. White boxes show overlapping predictions in the first three examples; colored boxes show differing predictions in the last example.
    }
    \label{fig:visualization}
\end{figure}

As shown in Fig.~\ref{fig:visualization}, we compare paired outputs to assess consistency between reasoning and grounding. The first three examples illustrate \textbf{misalignment} in the baseline: it correctly localizes the target but describes a different object. Specifically, it describes \textit{tissue paper} as a \textit{sponge}, a \textit{tray} as a \textit{bucket}, and \textit{scissors} as a \textit{knife}. Rita retains these localizations while producing descriptions that agree with the localized targets. The last example shows a grounding error accompanied by an incorrect object label: the baseline predicts a \textit{ladder} with an incorrect bounding box, whereas Rita identifies and localizes the \textit{stool} needed to reach the shelf.

\subsection{Zero-shot Evaluation}



\begin{figure}[t]
  \centering
  \begin{subfigure}[b]{0.48\textwidth}
    \centering
    \includegraphics[width=.92\linewidth]{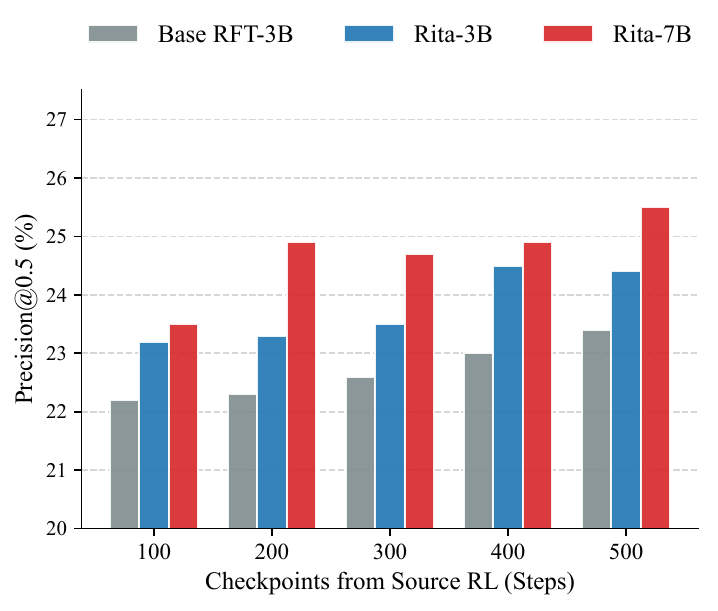}
  \end{subfigure}
  \hfill
  \begin{subfigure}[b]{0.48\textwidth}
    \centering
    \includegraphics[width=.92\linewidth]{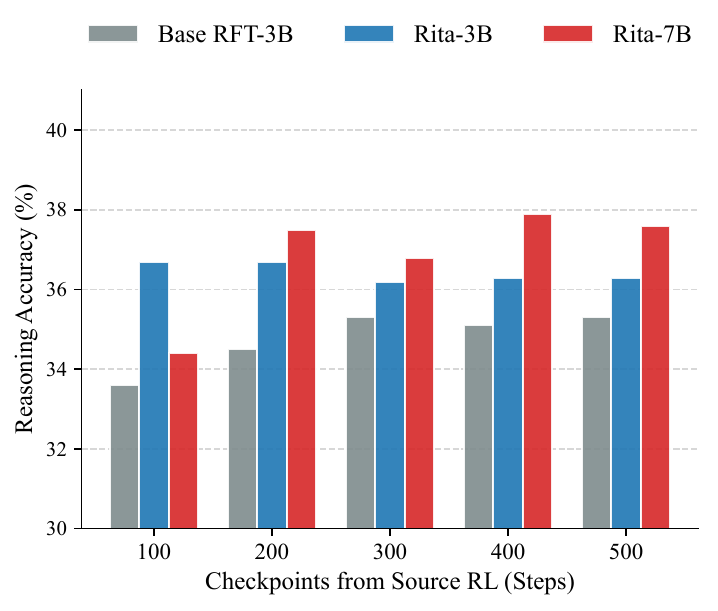}
  \end{subfigure}

  \caption{Zero-shot generalization across source RL training checkpoints on RefEgo-Int. We evaluate (a) \textbf{visual grounding precision} (P@0.5) and (b) \textbf{intention reasoning accuracy} using model checkpoints saved at different steps during RL training on the EgoIntention source dataset.}
  \label{fig:zeroshot_bars}

\end{figure}

To verify the generalization capabilities of our Rita framework, we evaluate the models on the challenging RefEgo-Int dataset in a zero-shot setting as shown in Fig.~\ref{fig:zeroshot_bars}.
Initially, the SFT baselines exhibit limited performance, with the 3B and 7B models achieving only 14.2\% and 18.0\% P@0.5, respectively. This can be attributed to the domain gap and inherent challenges in egocentric vision, such as motion blur, long-tail categories, and the requirement to reject queries when no target is present, a capability the original SFT models lack.
However, after applying our RL framework on the EgoIntention dataset, we observe substantial zero-shot improvements. Compared to the SFT baselines, our RL-tuned 3B and 7B models yield gains of +10.2\% and +7.5\% in P@0.5 respectively.
Furthermore, our method incorporating thinking and consistency rewards consistently outperforms the Base RFT model (IoU and format rewards only). At step 500, our 3B model achieves a 1.0\% lead in intention reasoning accuracy over the baseline, while the 7B variant achieves a peak reasoning accuracy of 37.6\%, demonstrating that our framework effectively fosters robust grounding-reasoning alignment that generalizes beyond the training distribution.

\section{Conclusion}


In this paper, we identify a mismatch between incorrect thinking processes and correct bounding boxes in visual intention grounding.
To address this issue, we propose Rita, which reinforces reasoning traces based on their causal contribution to predicting the correct answer.
Beyond the proposed probability-based rewards, we introduce a difficulty- and informativeness-aware data filtering strategy tailored to visual intention grounding, which focuses GRPO training on informative easy-to-medium samples.
Extensive experiments on EgoIntention and RefEgo-Int show that our Rita framework consistently improves both grounding accuracy and intention reasoning, achieving state-of-the-art performance over strong Qwen2.5-VL baselines.
We hope our framework will inspire future work toward more thought--answer consistent VLMs.



\section*{AI Usage Statement}
We used AI tools to assist in drafting sections of this paper and to polish its wording, grammar, and clarity.
Teacher-generated training traces used for baseline reproduction are described in Section 4.1 and Appendix C.
The authors take responsibility for the final content.

\bibliographystyle{iclr2027_conference}
\bibliography{main}

\clearpage
\appendix


This supplementary material provides the construction of the RefEgo-Int dataset (Appendix~\ref{sec:refego}), rollout statistics of the difficulty bins that motivate our data filtering (Appendix~\ref{sec:bins}), the thinking-drift analysis behind the statistics in the Introduction (Appendix~\ref{sec:drift_analysis}), ablations on the IoU margin of the consistency reward and on reference-answer overwrite (Appendix~\ref{sec:ablations}), the prompt template (Appendix~\ref{sec:prompt}), and a discussion of limitations and future work (Appendix~\ref{sec:limitations}).


\section{RefEgo-Int Dataset Construction}
\label{sec:refego}

RefEgo-Int is an image-based counterpart of RefEgo~\citep{kurita_refego_2023} that both preserves the original dataset's challenge distribution and augments each sample with a human intention sentence expressing the need for the target object, hence the suffix ``Int''. In total, we extract 1,317 images from RefEgo to construct RefEgo-Int.

\begin{figure}[h]
    \centering
    \includegraphics[width=1.0\linewidth]{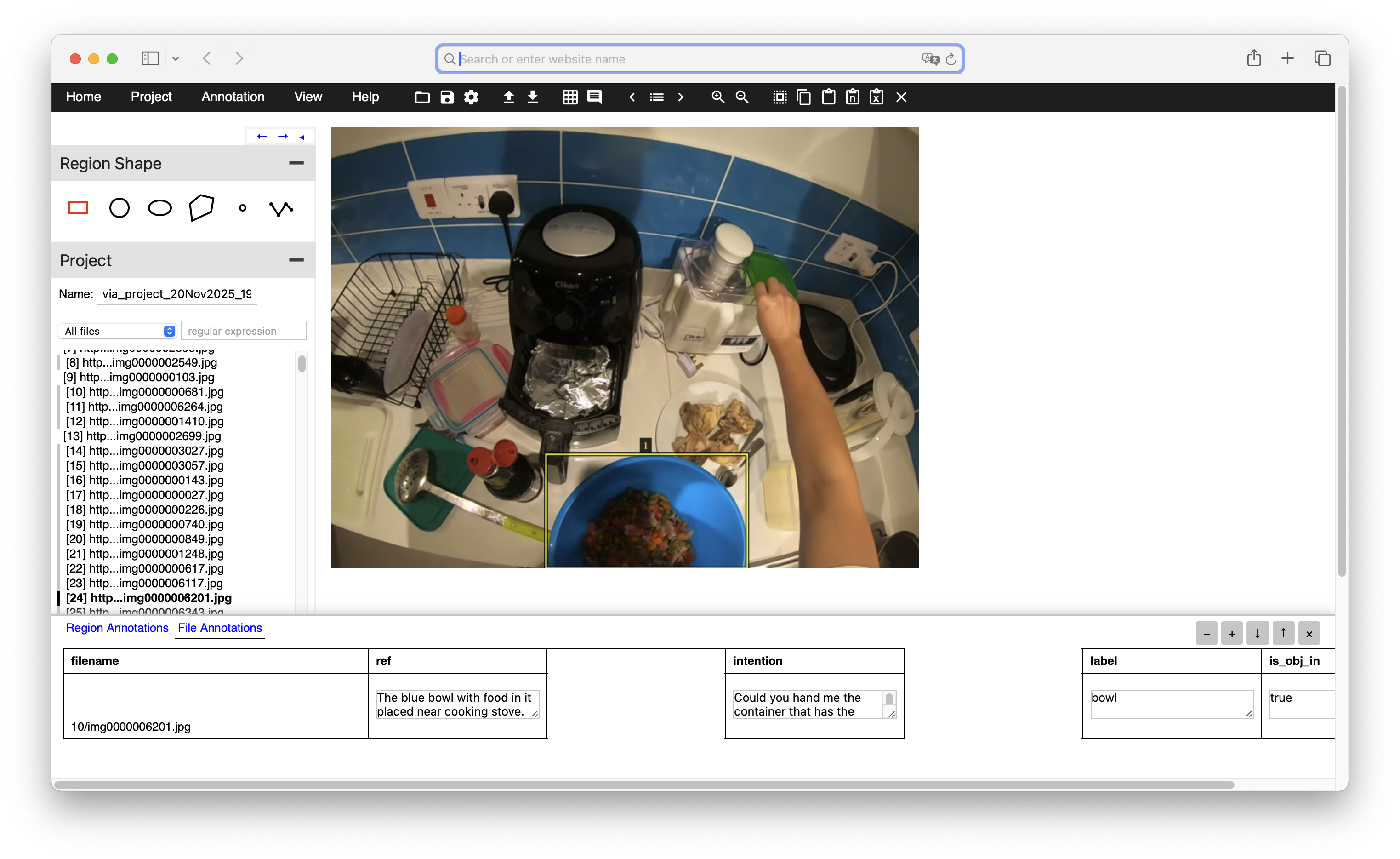}
    \caption{Screenshot of the VGG Image Annotator interface used to verify intention sentences and correct bounding boxes.}
    \label{fig:screenshot}
\end{figure}

\textbf{RefEgo$\rightarrow$RefEgo-Int (one-frame-per-clip).} To create an image version that faithfully reflects the video dataset's difficulty profile, we select exactly one annotated frame per clip while preserving the original $2{\times}2{\times}2$ challenge distribution over camera motion (stationary or moved), the number of referred objects (unique or multiple), and target visibility (visible or not). Concretely, we compute the annotation-level proportions of these 8 bins in RefEgo and perform stratified one-per-clip selection to match them in RefEgo-Int; when multiple frames qualify within a clip, we prefer the earliest frame.
This yields an image dataset that closely mirrors the video set's challenge mix (Table~\ref{tab:refego_image_creation}) while retaining a single, clean frame per clip for image-only evaluation.
We further annotate each RefEgo-Int sample with an intention sentence, following the annotation pipeline of EgoIntention~\citep{Sun_2025_ICCV}. Fig.~\ref{fig:screenshot} illustrates the procedure using VGG Image Annotator (VIA) ~\citep{dutta2019vgg} for checking intention sentences and collecting and correcting bounding box annotations.

\begin{table}[t]
    \centering
    \caption{Challenge distribution of RefEgo (annotation level) and RefEgo-Int (one frame per clip) over camera motion, number of referred objects, and target visibility.}
    \label{tab:refego_image_creation}
    \begin{tabular}{lllrr}
        \toprule
        Camera & Referred objects & Target visible & RefEgo & RefEgo-Int \\
        \midrule
        Moved & Multiple & No & 1264 (2.00\%)   &  27 (2.05\%)  \\
        Moved & Multiple & Yes & 3796 (6.01\%)   &  85 (6.45\%)  \\
        Moved & Unique & No & 1290 (2.04\%)   &  26 (1.97\%)  \\
        Moved & Unique & Yes & 4470 (7.08\%)   &  79 (6.00\%)  \\
        Stationary & Multiple & No & 6858 (10.87\%)  & 154 (11.69\%) \\
        Stationary & Multiple & Yes & 20762 (32.89\%) & 489 (37.13\%) \\
        Stationary & Unique & No & 6080 (9.63\%)   & 113 (8.58\%)  \\
        Stationary & Unique & Yes & 18600 (29.47\%) & 344 (26.12\%) \\
        \bottomrule
    \end{tabular}
\end{table}


\section{Statistical Characteristics of Sample Difficulty Bins}
\label{sec:bins}

\begin{table}[t]
\centering
\caption{\textbf{IoU rollout statistics across difficulty bins.}
We bin samples by the number of correct rollouts ($n_{\text{correct}}$ out of 8; IoU$\ge0.5$).
For each bin, we report the number of samples, the mean $n_{\text{correct}}$, the mean of per-sample mean IoU, the mean of per-sample IoU variance, and the mean fraction of zero-IoU rollouts (averaged per sample).}
\label{tab:iou_stats_bins}
\begin{tabular}{lrrrrr}
\toprule
\textbf{Bin} & \textbf{\#Samples} & $\boldsymbol{\mathbb{E}[n_{\text{correct}}]}$ & $\boldsymbol{\mathbb{E}[\mu_{\text{IoU}}]}$ & $\boldsymbol{\mathbb{E}[\sigma^2_{\text{IoU}}]}$ & $\boldsymbol{\mathbb{E}[\text{Frac}(\text{IoU}{=}0)]}$ \\
\midrule
Easy  & 9757 & 7.583 & 0.851 & 0.027 & 0.023 \\
Medium& 1988 & 4.118 & 0.479 & 0.133 & 0.240 \\
Hard  & 1536 & 1.411 & 0.213 & 0.078 & 0.452 \\
XHard & 2346 & 0.000 & 0.054 & 0.004 & 0.676 \\
\bottomrule
\end{tabular}
\end{table}

We analyze the rollout statistics on the context split of the EgoIntention training set to clarify why our data selection favors easy-to-medium samples for grounding RL (see Table~\ref{tab:iou_stats_bins}).
Easy samples provide clean but partially saturated supervision: they have high $\mathbb{E}[\mu_{\text{IoU}}]$ (0.851) and a very low zero-IoU rate (0.023), yet their small within-sample variance $\mathbb{E}[\sigma^2_{\text{IoU}}]$ (0.027) suggests limited outcome diversity across rollouts, which reduces the relative comparison signal exploited by GRPO updates. Medium samples strike a better balance between correctness and informativeness, achieving non-trivial localization quality ($\mathbb{E}[\mu_{\text{IoU}}]=0.479$) while exhibiting the largest variance (0.133), meaning that some rollouts succeed and others fail for the same input, yielding a stronger and more stable learning signal. In contrast, hard and extra-hard samples are dominated by failure rollouts: mean IoU drops to 0.213/0.054 and the fraction of zero-IoU rollouts rises to 0.452/0.676, making advantages weak or unstable; notably, XHard also has extremely low variance (0.004), indicating uniformly poor rollouts with little contrast to guide improvement. Overall, these statistics support training on easy-to-medium samples while filtering out very hard cases that mostly produce near-zero rewards. The XHard bin contains exactly the samples without any correct rollout ($d_i=0$ in Section~\ref{3.3}), which our filtering discards.

\section{Analysis of Thinking Drift under IoU-only RL}
\label{sec:drift_analysis}

This section quantifies the thinking drift that IoU-only RL introduces, as reported in the Introduction.
We analyze the UniVG-R1~\citep{bai_univg-r1_2025} re-implementation described in Section~\ref{sec:experiments}.
Two details differ from the original recipe: the Qwen2.5-VL-72B-Instruct teacher is conditioned on the ground-truth box and category, and it generates a single trace per sample instead of selecting the best of three candidates.
We compare the model after the chain-of-thought cold start with the model after 500 RL steps, sample by sample, on the 10{,}000 test samples of the \textit{Context} and \textit{Uncommon} splits.
A sample drifts if its predicted box is correct (IoU $\ge 0.5$) but its predicted object category differs from the ground truth.
As for Acc$_{\text{think}}$, the predicted category serves as a proxy for the target of the reasoning; we do not judge the \texttt{<think>} text itself.

\begin{table}[h]
\centering
\caption{\textbf{Thinking drift of UniVG-R1 before and after IoU-only RL} on EgoIntention (5{,}000 test samples per split). Drift is the percentage of samples whose box is correct (IoU $\ge 0.5$) but whose predicted category is wrong; the last row normalizes drift by the number of correct boxes.}
\label{tab:drift}
\resizebox{\linewidth}{!}{%
\begin{tabular}{lcc}
\toprule
 & Before RL (CoT cold start) & After IoU-only RL (step 500) \\
\midrule
P@0.5 (\%), Context / Uncommon / Overall & 64.90 / 57.82 / 61.36 & 65.86 / 59.60 / 62.73 \\
Drift (\% of samples), Context / Uncommon / Overall & 3.54 / 4.48 / 4.01 & 5.04 / 4.92 / 4.98 \\
Drift among correct boxes (\%), Overall & 6.54 & 7.94 \\
\bottomrule
\end{tabular}%
}
\end{table}

As shown in Table~\ref{tab:drift}, IoU-only RL raises P@0.5 by 1.4 points, while overall drift rises from 4.01\% to 4.98\% of samples.
The paired analysis locates this increase: RL newly corrects 539 boxes (and breaks 402), and 85 of the 539 newly corrected boxes (15.8\%) come with a wrong category.
This rate is $2.4\times$ the drift rate among correct boxes before RL (6.54\%), so the grounding gains of RL are disproportionately unsupported by a correct target.
The increase is statistically significant: 229 samples drift only after RL and 132 only before RL (exact McNemar test, $p=3.7\times10^{-7}$).
Because it is measured after the chain-of-thought cold start, it also shows that supervised reasoning traces do not prevent drift from emerging during RL.

\section{Additional Ablation Study}
\label{sec:ablations}

\subsection{IoU Margin of the Consistency Reward}
\label{sec:delta_ablation}

\begin{table}[h]
\centering
\caption{\textbf{Ablation on the IoU margin $\delta$ of the consistency reward.}
We report P@0.5 / Acc$_{\text{think}}$ / Acc$_{\text{align}}$ (\%) on the EgoIntention \textit{Context} and \textit{Uncommon} splits at different RL update steps. Both runs use Qwen2.5-VL-3B-Instruct and the full Rita reward; only $\delta$ differs. The main paper uses $\delta=0.1$.}
\label{tab:delta_ablation}
\resizebox{\linewidth}{!}{%
\begin{tabular}{c|cc|cc}
\toprule
\multirow{2}{*}{\textbf{Step}} & \multicolumn{2}{c|}{\textbf{Context}} & \multicolumn{2}{c}{\textbf{Uncommon}} \\
\cmidrule(lr){2-3} \cmidrule(l){4-5}
 & $\delta=0.1$ & $\delta=0.5$ & $\delta=0.1$ & $\delta=0.5$ \\
\midrule
100 & 66.3 / 83.8 / 61.2 & 66.0 / 84.6 / 61.5 & 60.0 / 71.1 / 53.1 & 60.9 / 73.3 / 54.9 \\
200 & 66.6 / 84.1 / 61.6 & 66.6 / 84.4 / 61.7 & 61.2 / 73.3 / 55.2 & 60.5 / 72.7 / 54.3 \\
300 & 66.8 / 84.0 / 61.7 & 67.4 / 86.2 / 62.8 & 61.2 / 72.7 / 54.8 & 62.2 / 74.2 / 55.7 \\
400 & 67.6 / 86.7 / 63.6 & 67.4 / 85.7 / 62.5 & 62.4 / 75.2 / 56.8 & 62.2 / 74.1 / 55.9 \\
\rowcolor{gray!8}
500 & \textbf{67.9} / \textbf{86.6} / \textbf{63.7} & 67.4 / 85.6 / 62.6 & \textbf{62.4} / \textbf{75.0} / \textbf{56.5} & \textbf{62.4} / 74.5 / 56.0 \\
\bottomrule
\end{tabular}%
}
\end{table}

The consistency reward in Eq.~\ref{eq:consistency_reward} rewards a rollout whose predicted box overlaps the target by more than $\delta$ and penalizes it otherwise, in proportion to its thinking score.
Table~\ref{tab:delta_ablation} compares our default margin $\delta=0.1$ with the standard grounding threshold $\delta=0.5$.
With $\delta=0.5$, the model is ahead on most metrics at steps 100 and 300.
With $\delta=0.1$, the model continues to improve and reaches the best final results: at step 500 it outperforms $\delta=0.5$ on all three metrics of the \textit{Context} split (+0.5 P@0.5, +1.0 Acc$_{\text{think}}$, +1.1 Acc$_{\text{align}}$) and on both reasoning metrics of the \textit{Uncommon} split (+0.5 each), with equal P@0.5.
A plausible explanation is that $\delta=0.5$ also penalizes rollouts whose boxes partially overlap the target, even when their traces point to the correct object, whereas $\delta=0.1$ restricts the penalty to boxes that largely miss the target, which better matches the thinking-drift failure the consistency reward is designed to suppress.



\subsection{Necessity of Reference-Answer Overwrite}
\label{sec:overwrite_ablation}

\begin{table}[h]
    \centering
    \caption{
        \textbf{Ablation on the effect of reference-answer overwrite for thinking reward.}
        We report Acc$_{\text{think}}$ (\%) on the EgoIntention \textit{Context} split at different RL update steps. \textit{Base RFT} uses only the IoU and format rewards. Without overwrite, the thinking reward does not improve over Base RFT.
    }

    \label{tab:logit}
    \centering
\resizebox{0.75\linewidth}{!}{%
    \begin{tabular}{l cccc >{\columncolor{gray!8}}c}
        \toprule
        Method & 100 & 200 & 300 & 400 & 500 \\
        \midrule
        Base RFT ($R_{\text{IoU}} + R_{\text{fmt}}$) & 82.2 & 81.0 & 81.9 & 83.0 & 83.7 \\
        \midrule
        + $R_{\text{think}}$ w/o reference-answer overwrite & 83.7 & 83.1 & 83.7 & 82.2 & 83.1 \\
        + $R_{\text{think}}$ w/ reference-answer overwrite & 83.8 & 83.6 & 83.7 & 84.2 & \textbf{85.1} \\
        \bottomrule
    \end{tabular}%
    }

\end{table}

We compare two variants of the thinking reward: one scores the model's own sampled answer (\textit{w/o reference-answer overwrite}), and the other replaces the answer tokens with the reference answer $A^\star$ as in Eq.~\ref{eq:answer_loglikelihood} (\textit{w/ reference-answer overwrite}).
As shown in Table~\ref{tab:logit}, at step 500 scoring the sampled answer lowers Acc$_{\text{think}}$ by 0.6 points relative to Base RFT, whereas scoring the reference answer raises it by 1.4 points, a 2.0-point gain over the variant without overwrite.
This matches the explanation in Section~\ref{3.2}: after a reasoning trace, the likelihood of the model's own answer mainly reflects its consistency with the trace rather than its correctness~\citep{welch2026cost}, so it also rewards confident incorrect predictions, whereas a fixed reference answer ties the reward to whether the trace supports the correct target.

\section{Prompt Template}
\label{sec:prompt}

Our RL training and evaluation use the referring-expression prompt of VLM-R1~\citep{shen_vlm-r1_2025} without modification.
The system turn is the Qwen default, and the intention sentence is inserted verbatim.

\noindent\fbox{\begin{minipage}{0.96\linewidth}\ttfamily\small\raggedright
{\normalfont\bfseries System:} You are a helpful assistant.\\[2pt]
{\normalfont\bfseries User:} <image> Please provide the bounding box coordinate of the region this sentence describes: \{intention sentence\} First output the thinking process in <think> </think> tags and then output the final answer in <answer> </answer> tags. Output the final answer in JSON format.
\end{minipage}}

\smallskip\noindent
The prompt does not request an object label.
The label in the answer is inherited from the supervised finetuning stage, whose targets follow the native Qwen2.5-VL grounding format, \texttt{[\{"bbox\_2d": [x1, y1, x2, y2], "label": "telephone"\}]}, inside a fenced JSON block.
Acc$_{\text{think}}$ and Acc$_{\text{align}}$ are scored on this label.

\section{Limitations and Future Work}
\label{sec:limitations}

\paragraph{Limitations.}
While \textit{Rita} encourages thought--answer consistency without reasoning labels, it introduces additional training overhead. Specifically, calculating $R_{\text{think}}$ requires a secondary forward pass where predicted answers are overwritten with reference-answer tokens ($A^\star$) to extract logits. This reprocessing step increases computational cost and memory usage compared to standard RL pipelines.

\paragraph{Future Work.}
We aim to extend our reward into a multimodal version that explicitly verifies visual grounding. Current rewards implicitly consider images through joint embeddings, but do not ensure the thinking trace actually leverages specific scene cues. To achieve a vision-dependent reward, we propose a thinking-guided selection mechanism: conditioning the reward on an attended or masked version of the image derived from the reasoning content. By filtering visual features based on the objects or relations mentioned in the trace, the reward would better reflect if the model's logic is grounded in actual visual evidence rather than dataset biases, further enhancing robustness and interpretability.

\end{document}